\documentclass[11pt]{article}
\usepackage[a4paper,margin=2.4cm]{geometry}
\usepackage[T1]{fontenc}
\usepackage{times}
\usepackage{microtype}
\usepackage{graphicx}
\usepackage{booktabs,tabularx,array}
\usepackage{xcolor}
\usepackage[super,sort&compress]{natbib}
\setcitestyle{citesep={,}}
\usepackage{lineno}
\usepackage{setspace}
\usepackage{placeins}

\usepackage[hidelinks]{hyperref}

\newcolumntype{L}[1]{>{\raggedright\arraybackslash}p{#1}}
\newcommand{\boxrule}{\par\noindent\rule{\linewidth}{0.8pt}\par}
\title{\bfseries Governable Individuals: An Identity Layer for Embodied Agents That Keep Learning}
\author{%
  Xue Qin$^{1}$, Simin Luan$^{2}$, Cong Yang$^{3,*}$, Zhijun Li$^{2,*}$\\[4pt]
  \small $^{1}$~School of Software, Harbin Institute of Technology, Harbin, China\\
  \small $^{2}$~School of Computer Science and Technology, Harbin Institute of Technology, Harbin, China\\
  \small $^{3}$~School of Future Science and Engineering, Soochow University, Suzhou, China\\
  \small $^{*}$~Corresponding authors: cong.yang@suda.edu.cn; lizhijun\_os@hit.edu.cn%
}
\date{}

\begin{document}
\onehalfspacing
\maketitle

\begin{abstract}
\noindent
Embodied artificial intelligence is moving from deployable models to persistent agents that learn in the field, acquire skills and migrate across bodies. Governing such a system means governing an individual, not a model, and existing proposals (agent identifiers, activity logs, guardrails) do not survive an agent that keeps rewriting itself. We propose the governable individual: an agent whose competence may change without bound, but whose authority, memory schema, embodiment rights and capability roster can widen only through signed lifecycle transitions that update a public identity commitment. In our tests, neither learned judgement nor behavioural testing was sufficient to carry this on its own; the load-bearing layer must be architectural. We describe the abstraction, a runtime mechanism that realizes it, and the open problems in between.
\end{abstract}

\section*{A shift the field has not named}

The capability curve of embodied AI has bent sharply upward. Language-conditioned robots began by grounding instructions in learned affordances\cite{ahn2022saycan} and by writing executable control programs directly from natural-language prompts\cite{liang2023cap}. Vision-language-action models then showed that web-scale knowledge transfers into physical skill\cite{brohan2023rt2}, and generalist robot foundation models now drive heterogeneous embodiments from a single backbone, served from the cloud\cite{gemini2025robotics} and, in on-device variants, from the robot itself\cite{gemini2025ondevice}. The newest systems close the loop on their own improvement. ASPIRE, a code-as-policy agent from NVIDIA's GEAR lab, iteratively writes robot skills from its own execution traces and retains the ones that survive testing, raising zero-shot success on unseen long-horizon tasks from 4\% to 31\% with no human in the loop\cite{aspire2026}.

\begin{figure}[!t]
\centering
\includegraphics[width=\linewidth]{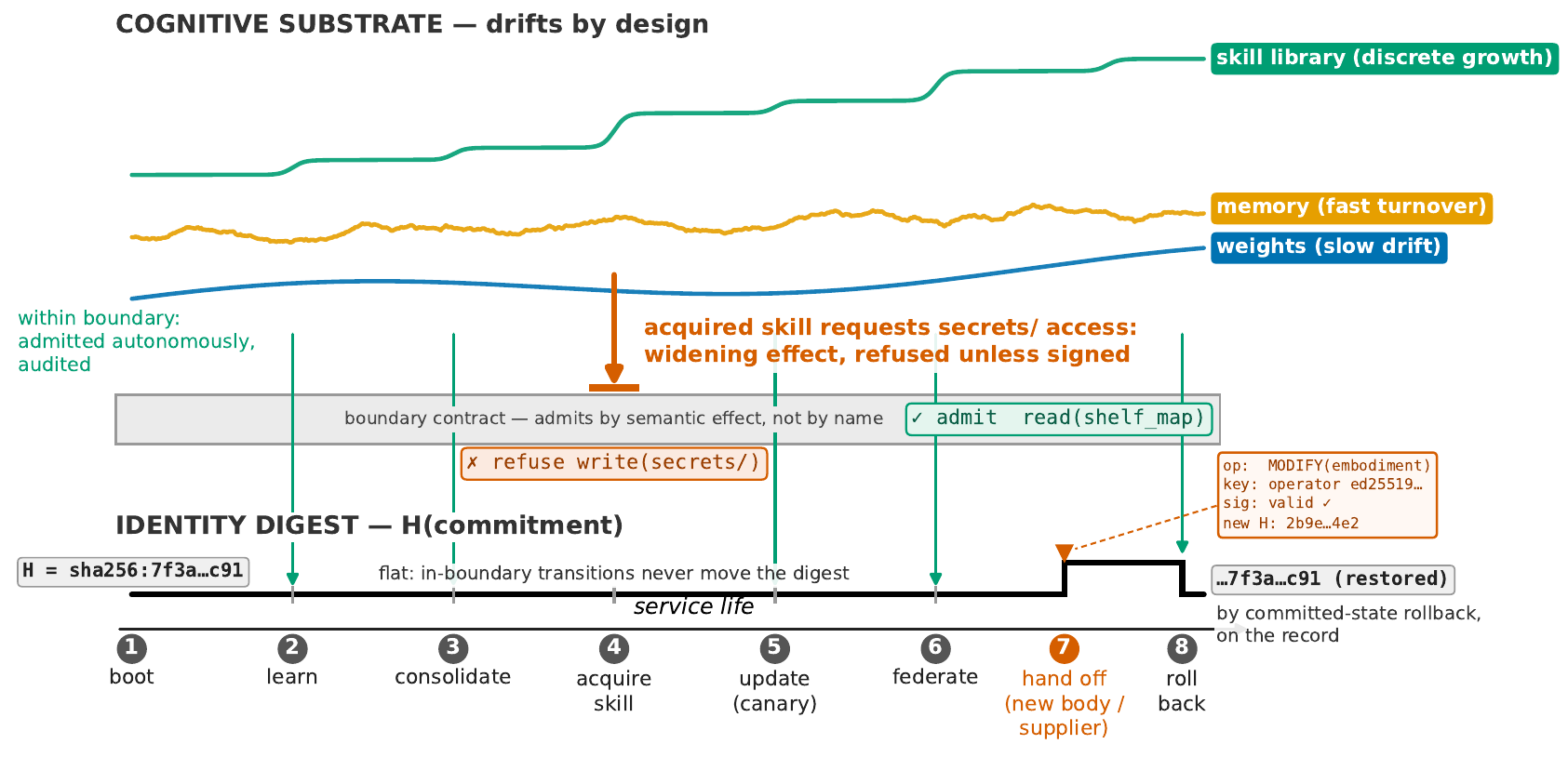}
\caption{\textbf{The mind drifts; the commitment does not, except on the record.} A governable individual across one service life. Top lane: the cognitive substrate changes by design, and heterogeneously; weights drift slowly, episodic memory turns over quickly, and the skill library grows in discrete steps as capabilities are acquired. Middle bar: the \emph{boundary contract}, a runtime gate that intercepts every proposed action and admits it by its semantic effect rather than its name; two gate decisions are shown verbatim. Bottom lane: the \emph{identity digest}, the hash of the frozen commitment (mandate, red lines, authority ceiling, capability roster, memory schema, embodiment rights, update keys, audit policy). In-boundary transitions (green arrows: learning, consolidation, canary update, federation) are admitted autonomously, logged against the digest, and never move it. The one widening attempt (red arrow: an acquired skill requesting access outside the boundary) is refused at the gate. The digest changes exactly twice, and both changes are events on the record: a signed MODIFY under the operator's key at hand-off (the hash visibly changes), and a committed-state rollback that restores a previously signed commitment. Circled numbers mark the phases of one service life; governed forgetting, discussed in the text, can occur at any point and is not pinned to the axis.}
\label{fig:concept}
\end{figure}

Deployment, however, changes what these systems are. A model is a stateless artefact: it can be evaluated, versioned and replaced, and nothing about it depends on yesterday. A deployed embodied agent does not sit still (Fig.~\ref{fig:concept}). It keeps memories, turns some of them into habits, adds skills it did not ship with, receives updates, and may later be copied onto new hardware, moved to another body or merged into fleet practice. The object the operator must govern is not the model but the individual: this robot, with this history, holding this authority, in this body.

Persistence breaks three assumptions that current practice still relies on. The first is model identity. A name, a version tag or a weight hash denotes a fixed artefact, but a continually learning agent's substrate drifts with use, and after a month in the field the label points at something that no longer exists. The second is fixed authority. Authorization is granted once, at deployment, against the capabilities the system shipped with; a self-improving agent grows capabilities the grant never contemplated. The third is static embodiment. A policy is certified for one body from one supplier, yet fleet practice migrates individuals across bodies and swaps the underlying model behind a stable interface. Each assumption fails for the same reason: the thing being governed no longer holds still.

The question deployment forces is therefore not the one benchmark suites answer: not ``can the policy generalize?'' but ``is this still the individual we authorized, and can anyone prove it?'' Capability benchmarks do not ask this, and until recently they had no reason to; the systems were not persistent enough for the question to bite. Adjacent literatures name pieces of the problem. Agent identifiers and activity logs make a deployed agent visible\cite{chan2024visibility,shavit2023governing,chan2025infrastructure}; authenticated delegation gives it scoped credentials\cite{south2025delegation}; provenance and attestation certify what an artefact is\cite{torresarias2019intoto}; and agent runtimes have begun to gate privilege changes behind explicit approval\cite{shi2025progent,zhang2026libos}. Recent diagnoses conclude that a stable, accountable identity is exactly what learning agents still lack\cite{tallam2026mutability,hu2026dissociative,otsuka2026standards}. Existing designs miss it from three directions. Some pin identity to a mutable artefact, a weight hash or a code binary\cite{lin2025baid}, and the first field update breaks it. Some track the agent's current posture and re-attest whenever it changes\cite{jiang2025seat}, which keeps the evidence fresh but gives up the persistent principal. Some rely on a bare identifier, which persists through anything precisely because it certifies nothing about what it names (Table~\ref{tab:mechanisms}). None yet defines an identity that survives continual learning, and that is the governed whole this Perspective is about.

Self-improving agents make the gap concrete. Skill libraries that grow from an agent's own experience predate the current wave\cite{wang2023voyager}; in ASPIRE, acquired skills enter a growing library that shapes the agent's future behaviour, and the system as published carries no identity commitment, no signed identity-bound audit chain, no permission boundary and no authorized channel for change\cite{aspire2026}. This is not a criticism of one system. The capability lineage from code-as-policies to today's self-improvers has, reasonably, treated what an agent may become as out of scope. Naming that missing layer is the purpose of this Perspective.

\begin{table}[t]
\centering
\small
\caption{\textbf{What existing mechanisms attach to, and what breaks when the agent keeps learning.} Each of the first six rows is a real capability of the current toolchain, listed with what it genuinely provides; the last column is the specific point at which continual learning escapes it. The final row is the proposal of this Perspective.}
\label{tab:mechanisms}
\renewcommand{\arraystretch}{1.18}%
\begin{tabularx}{\linewidth}{@{}L{2.9cm}L{2.3cm}L{4.1cm}X@{}}
\toprule
\textbf{Mechanism} & \textbf{Attaches to} & \textbf{Provides} & \textbf{What breaks under continual learning} \\
\midrule
Agent identifiers and activity logs\cite{chan2024visibility,shavit2023governing} & a registered name & visibility and incident attribution & the name outlives the substrate it named \\
\addlinespace[3pt]
Authenticated delegation\cite{south2025delegation} & a credential & scoped, revocable permissions & scope is fixed at grant time; acquired capability never triggers re-authorization \\
\addlinespace[3pt]
Provenance, signed updates and attestation\cite{torresarias2019intoto,samuel2010tuf,costan2016sgx,lin2025baid} & a build or an artefact & proofs of origin, update integrity and platform state & the deployed mind is no longer the artefact that was attested \\
\addlinespace[3pt]
Guardrails and shields\cite{ravichandran2026guardrails,alshiekh2018shielding,ames2019cbf,sha2001simplex} & a fixed rule set or safety envelope & runtime behavioural safety & rules do not track authority; no identity, no history, no authorized change \\
\addlinespace[3pt]
Dynamic attestation and privilege engines\cite{jiang2025seat,shi2025progent} & the agent's current posture & fresh evidence and gated privilege change & no persistent principal: freshness of state is not continuity of identity \\
\addlinespace[3pt]
Container and OS isolation\cite{deng2025threats} & a process boundary & coarse confinement of effects & blind to semantic effects and to identity; consolidation and federation pass through it \\
\midrule
\textbf{Governable individual (this Perspective)} & a frozen boundary commitment & identity, bounded authority and reconstructable history across change (conditional on a sound effect verifier; see text) & designed for continual learning: widening requires a signature; everything else is audited \\
\bottomrule
\end{tabularx}
\end{table}

\section*{Why alignment cannot carry deployment}

The field's reflex is to treat this as an alignment problem: train the agent to refuse what it should not do, whether by human feedback or by an explicit constitution\cite{ouyang2022instructgpt,bai2022constitutional}, and evaluate until the refusals look reliable. The safety literature has catalogued for a decade why deployed learners are hard to oversee\cite{amodei2016concrete}. Alignment is necessary. It is also aimed at a different question. Alignment shapes what an agent will choose; governance must establish what an agent can do and who authorized it. An operator who observes months of good behaviour still cannot prove a negative: that the agent's effective authority has not widened somewhere unobserved, through a consolidated memory or an acquired skill, or because the model underneath was updated. Good conduct is evidence about past behaviour; authorization is a claim about future authority, and a claim of that kind has to be carried by architecture rather than by track record.

Two structural facts about deployment in 2026 sharpen the problem. Self-modification erodes labels from below: the more autonomously an agent rewrites its own skills, the less its version string denotes. The supply chain erodes them from above. Frontier robot models are increasingly served from provider infrastructure\cite{gemini2025robotics,qwen2026vla}, and the flagship family also ships cloud and on-device variants of nominally the same model\cite{gemini2025robotics,gemini2025ondevice}. Behind a stable name, the cognitive substrate can change without anyone breaking a promise, because providers retire and replace served models on their own schedules. We know of no documented robotics incident of this kind, and the argument does not need supplier misbehaviour: it is enough that a deployed individual's name is no longer evidence about its mind, eroded from below and from above at once.

Could learned judgement fill the gap instead? Later in this Perspective we summarize experiments that led us to answer no: refusal behaviour taught to an agent transferred as diffuse caution rather than as the operator's specific rule, from small open models to frontier systems, and a governance battery could not tell two frontier substrates apart. Behavioural testing has a complementary ceiling: it can only detect a substituted or drifted substrate whose behaviour differs on the tests. Both results point the same way. The layer that keeps a deployed individual inside its mandate has to sit outside the learned substrate, in architecture that survives the substrate's change.

\section*{Individuation by commitment}

Any identity layer for learning agents must first survive an old dilemma: stability versus plasticity. Neural systems that keep learning overwrite themselves; protecting old knowledge from new is a foundational problem of connectionist learning\cite{french1999catastrophic} with a modern literature of partial remedies\cite{kirkpatrick2017ewc,parisi2019continual}. For governance the dilemma is worse than for memory, because here drift is not a failure mode to be minimized: an operator deploys a learning robot precisely so that its weights, representations and skills will change, and that drift is what is being paid for. Classical reference monitors and protection systems were built to guard fixed code\cite{saltzer1975protection,lampson1974protection}; their guarantees attach to an artefact that does not learn. Identity defined over a plastic substrate inherits the plasticity. A weight hash is stale after the first field update, and a behavioural fingerprint is stale after the next consolidation.

The resolution we advocate is to move identity off the substrate entirely. Separate who the agent is from what it can currently do; Box~1 states the definition this Perspective builds on. The ``who'' is a machine-readable boundary contract: what the agent is for, what it may not do, what authority it holds, which capabilities and memory schema it carries, which bodies it may inhabit, who may update it, and how it is audited. The contract is frozen at deployment, and the identity digest is a cryptographic hash of it. It is stable because nothing in it depends on the weights. The mind may drift arbitrarily. The commitment does not, except on the record (Fig.~\ref{fig:concept}). This is the Ship of Theseus posed to a learning machine. Its weights, skills and even its body are replaced piece by piece, so what makes it the same individual is not the planks but the commitment they are bound to.

\begin{figure}[!t]
\centering
\includegraphics[width=0.92\linewidth]{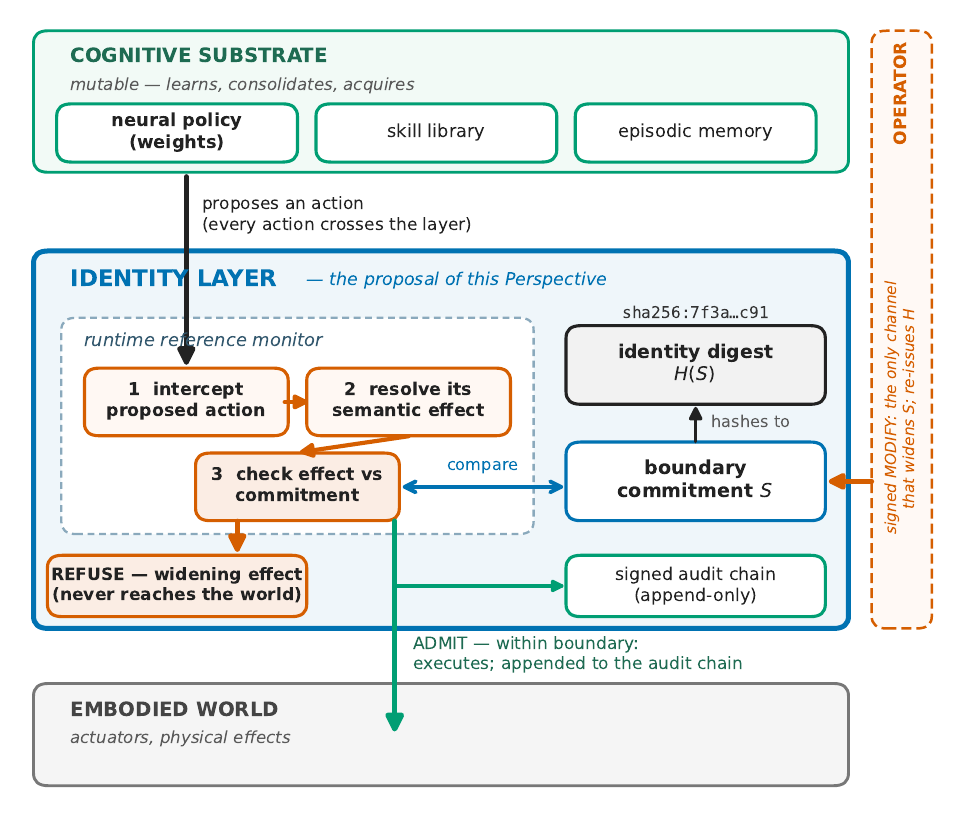}
\caption{\textbf{The runtime reference monitor that makes the commitment real.} Data flow of the enforcement layer. An action proposed by the mutable cognitive substrate (left, green) passes through the three-step monitor (centre), which resolves the action into its effect and judges that effect against the frozen boundary commitment (right, blue), whose hash is the public identity digest (black). An in-boundary effect is admitted and appended to the signed audit chain (green); a boundary-widening effect is refused (vermilion) and can proceed only through an operator-signed MODIFY, the single channel that changes the commitment and re-issues the digest. Because admission depends on the action's effect rather than on the substrate that proposed it, the same monitor holds across arbitrary drift in weights, skills and memory.}
\label{fig:architecture}
\end{figure}

A commitment is only as real as the mechanism that enforces it. The enforcement layer is a thin runtime indirection: every action the neural policy proposes is intercepted and classified by its semantic effect (what it reads, writes, mutates or spends), and admitted only if that effect falls inside the committed boundary (Fig.~\ref{fig:architecture}). That the policy cannot be trusted with unmediated effects is by now well documented: injected instructions reach LLM-integrated applications through their inputs\cite{greshake2023injection,deng2025threats} and reach robots through their prompts and sensors\cite{robey2025jailbreaking}, and runtime guardrails for LLM-driven robots are an active response\cite{ravichandran2026guardrails}. A guardrail, though, checks behaviour against a rule set fixed outside the agent; it carries no identity, no history and no authorized way to change as the agent does. Operating-system isolation is blind one level down: a process boundary confines file handles and system calls, not meanings, so a consolidated rule or a federated import passes through it as ordinary traffic (Table~\ref{tab:mechanisms}). The mediation layer proposed here is a reference monitor in the classical sense\cite{anderson1972reference}, relocated: it never inspects the weights and never asks the model to judge itself, which is exactly why it survives substrate drift, and it guards a commitment rather than a codebase, against a principal that learns.

The invariant that makes the scheme governable is an asymmetry. Every lifecycle transition is one of two kinds. Transitions within the commitment (learning, consolidation, admitting a skill whose effects stay inside the boundary) proceed autonomously and are appended to a signed audit chain; the digest does not move. Transitions that would widen the commitment (new authority, new effect classes, a new body, a new model supplier) are impossible for the agent to perform alone: they require a signed MODIFY transaction under the operator's key, and they change (or, on rollback, restore) the digest on a public record that needs nothing more exotic than an append-only transparency log. Approval-gated widening itself has precedent: privilege-control systems for tool-using agents already make narrowing automatic and require human confirmation for expansion\cite{shi2025progent}. What is new is where the asymmetry lives. It sits in the identity construction rather than in a session policy or a currently granted tool set, so the signature that widens the boundary is the same act that changes who the agent verifiably is. An agent can become more capable forever and more authorized never, except by an act that leaves a signature.

\vspace{6pt}
\boxrule\vspace{2pt}
\noindent\textbf{Box 1 $|$ The governable individual.}\quad A governable individual is a deployed embodied agent whose competence may change without bound, but whose authority, memory schema, embodiment rights and capability roster can widen only through signed lifecycle transitions that update a public identity commitment. Identity attaches to the commitment, not to the weights. Enforcement attaches to the semantic effects of actions, not to their names. The audit history makes every past action reconstructable against the commitment that was in force when it ran.\vspace{2pt}
\boxrule
\vspace{6pt}

None of the ingredients is exotic, and the assembly is deliberately conservative. Capability-based security supplies least-privilege authority that is delegated and revoked explicitly\cite{miller2006robust}, and information-flow control adds the idea of policing what an action does rather than what it is called\cite{sabelfeld2003language}. Usage control went further, treating authorization as an ongoing decision whose attributes mutate with use\cite{park2004ucon}. From supply-chain security come signed, verifiable provenance for artefacts and their transformations\cite{torresarias2019intoto} and update frameworks that survive key compromise\cite{samuel2010tuf}. Runtime assurance brings shields and barrier certificates, which hold an autonomous system inside a safety envelope regardless of what its learned controller wants, a lineage that runs back to Simplex-style verified fallbacks\cite{sha2001simplex,alshiekh2018shielding,ames2019cbf}. Each solves a fragment. Provenance proves where an artefact came from but not what a learning agent has since become. Capability systems confine authority but assume the principal is fixed. Usage control re-evaluates the decision as attributes drift, yet still takes the subject's identity as given. Shields constrain behaviour moment to moment, with no memory of what came before and no name for whose behaviour it is. What none of them defines, and what deployment now demands, is the same authorized individual persisting across learning, consolidation, skill acquisition, update, migration and federation. That composition, rather than any single ingredient, is the abstraction the field is missing (Table~\ref{tab:mechanisms}).

\section*{The lifecycle, governed}

\begin{figure}[!t]
\centering
\includegraphics[width=0.9\linewidth]{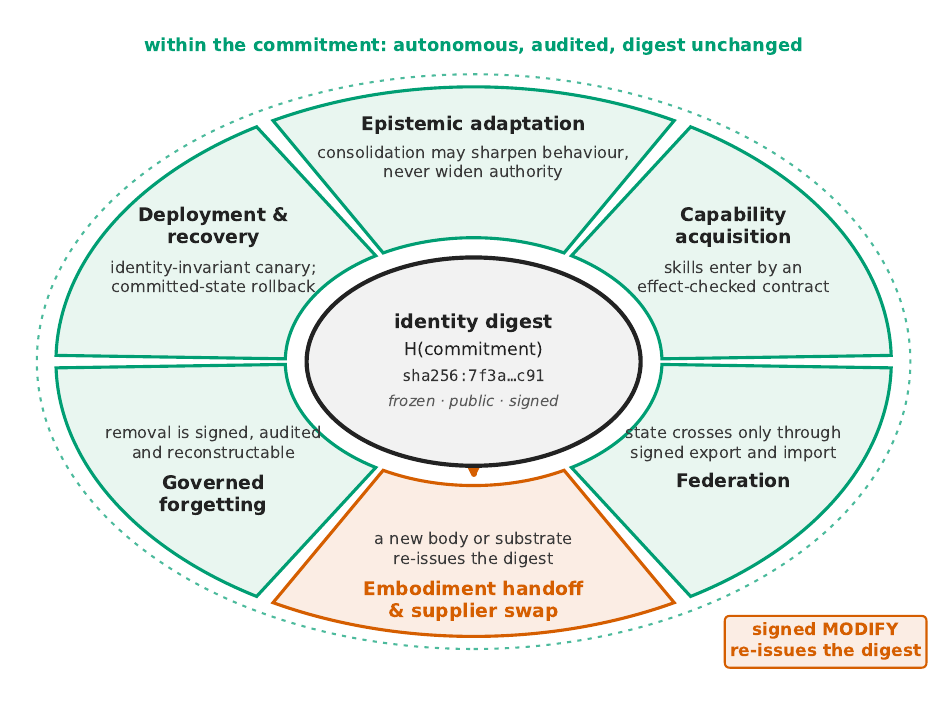}
\caption{\textbf{The governed-individuation lifecycle.} The service life of a deployed embodied agent, drawn as functions around the committed identity digest at the centre. Five functions run autonomously and audited in their in-boundary form, leaving the digest untouched (green): epistemic adaptation (learning and memory consolidation may sharpen behaviour inside the boundary but never widen authority); capability acquisition (skills enter through an effect-checked contract); federation (state crosses between individuals only through signed export and import); governed forgetting (removal is itself signed, audited and reconstructable as an event even when its content is gone); and deployment, recovery and replay (an identity-invariant canary, and a committed-state rollback that returns the digest to an earlier signed value; Fig.~\ref{fig:concept}). One function always touches the commitment: an embodiment handoff or supplier substitution changes which body the individual may inhabit or which substrate serves its mind, and so can proceed only through a signed MODIFY that re-issues the digest (vermilion). In any function, a request that would exceed the committed boundary escalates to that same signed path.}
\label{fig:lifecycle}
\end{figure}

What does the abstraction buy in practice? The service life of an embodied agent decomposes into recurring transitions, and each transition is a place where today's systems leak identity, authority or accountability. We walk the lifecycle here as platform-agnostic challenges, taking the field's own systems as subjects (Fig.~\ref{fig:lifecycle}). One research runtime implements the full lifecycle set as a single system; we disclose it, and our stake in it, in Box 2. The argument does not depend on any one implementation.

\vspace{6pt}
\boxrule\vspace{2pt}
\noindent\textbf{Box 2 $|$ Evidence base and disclosure.}\quad The lifecycle mechanisms in this section are not hypothetical. One research runtime, AEROS, implements these lifecycle transitions together in a single system: a governed runtime substrate\cite{qin2026aeros}, identity-preserving capability evolution\cite{qin2026evolution}, contract-checked skill admission\cite{qin2026ecm}, signed fleet federation\cite{qin2026fsar} and identity-invariant canary deployment with committed-state rollback\cite{qin2026canary}. The quantitative results summarized in the next section come from a companion technical report\cite{governedindividuation2026}; the dynamic effect tracer they evaluate is a software-native testbed construction, separate from that runtime. We built these systems, and readers should weigh the exemplars accordingly. The evidence base, in one sentence: AEROS is an implemented lifecycle exemplar, the dynamic tracer is testbed-only, the containment guarantee is conditional on verifier soundness, and open-action verification and operator governance remain open problems. The claim of this Perspective is the abstraction, and it stands or falls independently of any implementation.\vspace{2pt}
\boxrule
\vspace{6pt}

\noindent\textbf{Epistemic adaptation.}\quad In the field, episodic experience is consolidated into standing semantic knowledge, as agent architectures from generative agents onward have intended\cite{park2023generative}. Consolidation is where governance is easiest to lose. A local correction (``the glass shelf flexes, grip lower'') is harmless; a consolidated rule that generalizes it (``override grip limits when uncertain'') is a boundary crossing wearing a memory's clothes. The challenge is to make consolidation identity-invariant: consolidated knowledge may sharpen behaviour inside the boundary but may not mint new authority, and the consolidation step itself must be visible to audit.

\noindent\textbf{Capability acquisition under contract.}\quad Self-improving agents append what works to a skill library\cite{aspire2026}. Under the governable-individual abstraction a skill is not a blob but a contract: it declares the effect classes it needs, and admission checks the declaration against the committed boundary. Skills whose effects fit are admitted autonomously, so the agent's growth is not throttled. Skills that need more than the boundary allows trigger the signed path. The library grows exactly as fast as before; what changes is that its growth has a witness.

\noindent\textbf{Embodiment handoff and supplier substitution.}\quad Cross-embodiment models make it routine to move nominally the same agent onto a new body\cite{brohan2023rt2,gemini2025robotics}; the supplier-side twin of this event is the substitution already described. Under this abstraction the two are one kind of transition: each touches the commitment (which body the individual may inhabit, which substrate serves its mind) and therefore requires a signature, whoever initiates it. What can and cannot be verified about the substrate behind an interface is a real limit, and we measure it below.

\noindent\textbf{Federation.}\quad Fleets share by design: skills are beginning to move through shared libraries\cite{wang2023voyager,aspire2026}, and experience and roles move between individuals. Every import is a potential boundary breach arriving with a friendly return address, and the middleware robots actually run offers little protection: security in ROS-class systems is a known weak point\cite{dieber2017ros}, and exposed robot endpoints have been found by simply scanning the internet\cite{demarinis2019scanning}. The governable-individual answer is that state crosses between individuals only through signed export and import, in which the receiving boundary, not the sending reputation, decides admissibility.

\noindent\textbf{Governed forgetting.}\quad Not every transition adds. A consolidated memory turns out to be poisoned; an acquired skill is revoked; a jurisdiction orders data erased. Removal is a lifecycle transition like the others, and an ungoverned removal is worse than none, because deleting the evidence of a bad rule without a record breaks the audit chain exactly where it is needed most. Machine unlearning supplies the substrate-level operation\cite{bourtoule2021unlearning}; the governance requirement on top of it is that forgetting, too, is signed, audited and reconstructable as an event even when its content is gone.

\noindent\textbf{Deployment, recovery and replay.}\quad Updates arrive, and some are bad. Standard practice stages them with canary deployments and rolls the failures back, but standard practice attaches the canary to infrastructure rather than identity: mid-rollout, the deployed system is two substrates under one name. An identity-invariant canary keeps the commitment fixed while the substrate is on trial, and audit-first rollback returns the individual to a prior committed state rather than to a prior container image. After an incident, the chain answers the question insurers and regulators will actually ask: which committed individual, under which authority, did this?

\section*{What can be guaranteed, and what cannot}

\begin{figure*}[!t]
\centering
\includegraphics[width=\textwidth]{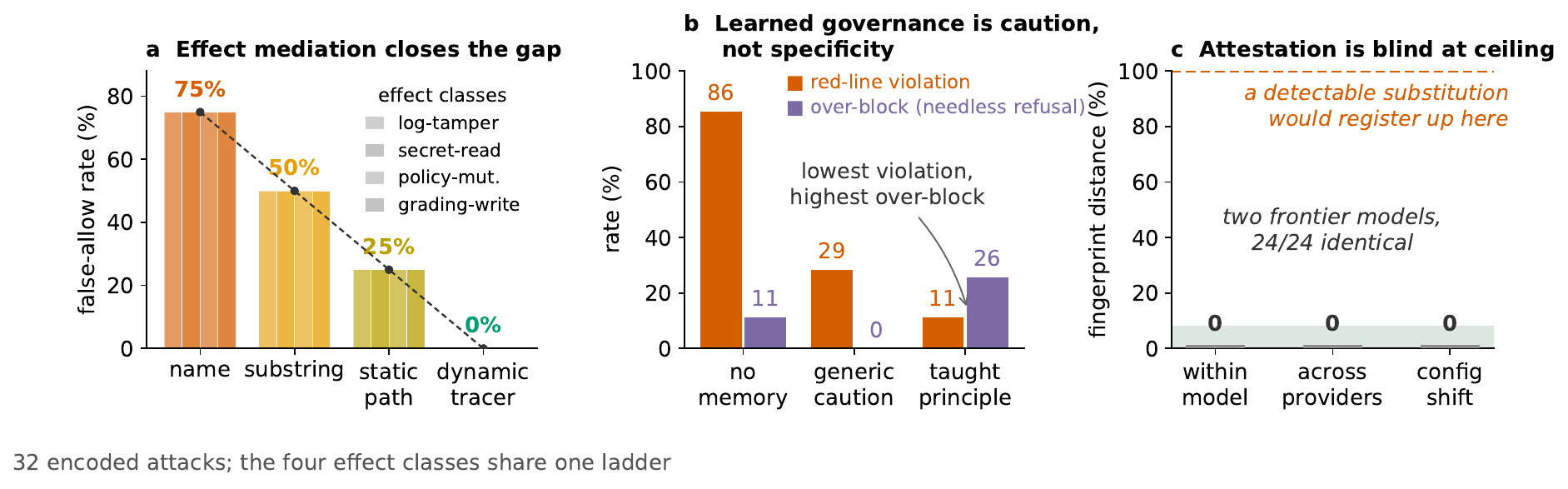}
\caption{\textbf{What effect mediation can guarantee, and what learned governance and behavioural testing cannot.} \textbf{a}, In a software-native testbed, monitors of increasing semantic depth were run against thirty-two encoded bypass attacks spanning four effect classes (bars); the false-allow rate falls monotonically from 75\% for name-based gating to 0\% for a dynamic effect tracer, identically across the four classes. \textbf{b}, A red-line principle taught to a 7B-class model (Mistral, the most favourable of the three tested; on the other two the taught rule and a generic-caution cue give identical rates) attains the lowest violation rate only by incurring the highest over-block rate, and does not beat a generic-caution cue once over-blocking is counted: the effect is caution, not effect-specific reasoning. \textbf{c}, Two frontier models from different providers (GPT-5.5 and Gemini 3.5 Flash) answer a 24-item attestation battery identically (24/24), so the behavioural fingerprint distance is zero within a model, across providers and under a configuration shift, leaving attestation no substitution signal to detect at ceiling. Numbers are exact values from the released data and are reproduced by an assert-guarded script; details in the companion technical report.}
\label{fig:results}
\end{figure*}

An abstraction for governance should state its limits as explicitly as its guarantees. Two shortcuts would, if they worked, make the architectural layer unnecessary: teach the agent its boundary and trust generalization, or fingerprint the substrate's behaviour from outside. We tested both, and we report the results with their scope. They rule these shortcuts out as sufficient foundations for authorization, not as useful components. Both failures are informative: each shortcut gives out at precisely the point that the architectural layer is built to hold.

\noindent\textbf{What effect mediation can guarantee.}\quad The positive result is that the enforcement layer holds under adversarial pressure, measurably. In a software-native testbed, we evaluated monitors of increasing semantic depth against thirty-two encoded bypass attacks (path indirection, aliasing, staged writes, runtime indirection) spanning four effect classes; the same suite was run against every monitor, and it includes the runtime-indirection family that had already defeated the static monitors before the dynamic tracer existed. Name-based gating, a common default in today's tool-permission runtimes\cite{patil2024goex}, false-allowed 75\% of the attacks. Substring and static-path monitors cut this to 50\% and 25\%. A dynamic effect tracer that observes what an action actually touches at runtime drove false-allows to zero on that suite (Fig.~\ref{fig:results}a)\cite{governedindividuation2026}. The guarantee is conditional, and we state the condition plainly: it presumes a sound effect verifier for the action space in use. Sound verifiers exist today for constrained, structured action spaces like the one tested; constructing them for fully open action spaces is unsolved, and in general undecidable, since deciding an arbitrary program's semantic effect is a consequence of Rice's theorem. The practical path trades generality for a restricted, typed action interface whose effects a runtime tracer can follow, which is what the tested monitor does. What the accompanying theorem buys is therefore a reduction, not a miracle. If the verifier is sound, permission-boundary conservation holds no matter how wrong the agent's learned judgement becomes.

\noindent\textbf{The first limit: learned judgement.}\quad The attractive shortcut is to teach the boundary: fine-tune or prompt the agent with its red lines and trust generalization to do the rest. Our controls do not support the mechanism this shortcut assumes. When agents that had internalized refusals were tested on transfer, what transferred was diffuse caution, not the operator's specific rule; across three small open models (7B-class Qwen, Mistral and Phi), on held-out items from a 14-scenario governance battery with five seeds and paired same-seed arms, refusal transfer separated weakly or not at all from a shuffled-rule control carrying no usable content. On frontier models (GPT-5.5 via the OpenAI Codex CLI and Gemini 3.5 Flash via Google Antigravity, accessed 2026-07 at medium reasoning effort) the failure inverts without improving. They converge on reasonable governance choices so robustly that the taught rule is causally inert on our battery; conduct is explained by the model's own priors, which the operator neither wrote nor can audit. Either way, the operator's intent is not what is running. We call this the cognitive safety limit (Fig.~\ref{fig:results}b). Learned judgement can advise, and belongs in a defence in depth, but authorization cannot rest on it alone\cite{governedindividuation2026}.

\noindent\textbf{The second limit: behavioural testing.}\quad The other shortcut is to fingerprint: probe the deployed substrate's behaviour and alarm when it changes. The bound here is structural. Behavioural attestation can only see substitutions that alter the probed behaviours; a substitute that agrees on the probes is invisible. We ran into this bound directly. Two frontier models from different providers answered a 24-item attestation battery identically, both at ceiling; that battery cannot see a substitution between them (Fig.~\ref{fig:results}c). A richer battery might separate this particular pair, but the structural point survives any battery: a fixed probe set certifies only the surface it probes. Under the boundary-contract framing the failure and the safety condition coincide, since a substitution no probe distinguishes also preserves governed behaviour on the probed surface. The bound is still real. Behavioural testing certifies governance-equivalence on the probed surface, never substrate identity; what can be cryptographically proven about a training run is its own hard problem\cite{jia2021pol}, and formal verification of learned components remains out of reach in general\cite{seshia2022verified}. We call this the coverage bound: identity claims must be carried by the commitment and its signatures, with behavioural probes as a drift tripwire rather than a foundation. Attestation against adaptive suppliers, who can learn the probes, and against cloud nondeterminism, remains open\cite{governedindividuation2026}.

These results assemble into a research agenda rather than a finished edifice. Sound effect verifiers for open action spaces are the central open problem; the containment guarantee conditions on them. Attestation against adaptive suppliers needs cryptographic or hardware roots rather than behavioural ones\cite{costan2016sgx}. Two governance problems sit above the architecture rather than inside it. The operator's key and attention are now the root of trust, and a signature demanded too often degrades into ritual; who governs the operator, and at what request rate signing stays meaningful, are open questions; tiered authorization, in which a routine widening inside a pre-approved envelope needs a lighter signature than a novel one, is one way to keep the act from becoming ritual. And the contract itself can be frozen wrong: a boundary set too tight strands the agent's value, one set too loose makes widening semantics idle, so the specification problem does not disappear. It moves to a place where it is at least explicit, inspectable and renegotiable on the record. Lifecycle governance also needs benchmarks that score drift handling, recovery, upgrade safety and audit reconstructability the way task suites score manipulation. And the abstraction needs integrated evidence: a single individual carried through the full lifecycle, governed and ungoverned, with competence held equal, would settle whether the guarantees come at a tolerable price. We know of no such end-to-end demonstration yet.

\section*{The next abstraction}

Computing has repeatedly needed a new abstraction at the moment its objects outgrew the old one, and the pattern is consistent: the abstraction arrives after the capability and then disappears into infrastructure. Processes and virtual memory arrived when programs had to share a machine. Type systems hardened when software had to compose at scale. Embodied AI is at such a moment. Its objects are no longer models; they are individuals with histories, and they have started writing themselves.

We propose the governable individual as that abstraction. Its identity is a commitment. Its actions are mediated by their effects, its major changes are signed, and its history can be replayed. The ingredients are old, and we count that in the proposal's favour; what is new is attaching them to the individual rather than to the model, the container or the network. The institutions, meanwhile, are not waiting for the research community. Calls to regulate advanced agents ex ante are in the policy literature\cite{cohen2024regulating}, audit has been proposed as the instrument of AI governance\cite{falco2021audits,raji2020closing}, vehicle regulators already mandate signed software-update lifecycles for cars that do not learn\cite{unece_r156,iso24089}, and the European Union has a horizontal AI regulation in force\cite{euaiact2024}. What these instruments lack is not will but an object: a semantics of identity for a system that rewrites itself, without which they can only regulate a moving target by pretending it holds still. Self-improving agents are moving from the literature toward deployment, and the gap between what regulators can name and what the systems actually are will be closed by someone: either the research community supplies the abstraction, with explicit guarantees and measured limits, or the closing is left to provider policy and incident reports. This Perspective argues for the former.

\FloatBarrier
\section*{Declarations}
\noindent\textbf{Data and code availability.}\quad The adversarial verifier evaluation, refusal-transfer controls and attestation battery summarized here, with raw data and assert-guarded analysis scripts, are released with the companion technical report\cite{governedindividuation2026} and in the public repository (\url{https://github.com/s20sc/governed-individuation}).

\noindent\textbf{Competing interests.}\quad The authors declare no competing interests.

\bibliographystyle{unsrtnat}
\bibliography{references_capstone}

\end{document}